\def\year{2021}\relax
\documentclass[letterpaper]{article} 
\usepackage{aaai21}  
\usepackage{times}  
\usepackage{helvet} 
\usepackage{courier}  
\usepackage[hyphens]{url}  
\usepackage{graphicx} 
\usepackage{natbib}  
\usepackage{caption} 
\usepackage[utf8]{inputenc}
\usepackage{comment}
\usepackage{amsmath,amssymb, mathtools} 
\usepackage{color} 

\def\bth{\bs{\theta}}

\def\bx{\bs{x}}

\def\bx{\bs{x}}

\def\bP{\bs{P}}

\def\bF{\bs{F}}
\def\bI{\bs{I}}

\def\bS{\bs{S}}

\def\miou{\text{mIoU}}
\newcommand{\ffigure}[1]{fig.~\ref{#1}}
\newcommand{\Figure}[1]{Fig.~\ref{#1}}
\newcommand{\Table}[1]{Table~\ref{#1}}
\newcommand{\Section}[1]{\S\ref{#1}}
\newcommand{\tableref}[1]{table~\ref{#1}}

\newcommand{\bs}{\boldsymbol}

\newcommand{\pith}{\pi_{\theta}}

\title{Embodied Visual Active Learning for Semantic Segmentation}

\author{\Large \textbf{David Nilsson\textsuperscript{\rm 1,2}\thanks{Work was partially performed during a Google internship.},  Aleksis Pirinen\textsuperscript{\rm 1}, Erik Gärtner\textsuperscript{\rm 1,2}\footnotemark[1], Cristian Sminchisescu\textsuperscript{\rm 1,2}}\\} 

\affiliations{
    \textsuperscript{\rm 1}Department of Mathematics, Faculty of Engineering, Lund University\\
    \textsuperscript{\rm 2}Google Research\\
    \{david.nilsson, aleksis.pirinen, erik.gartner, cristian.sminchisescu\}@math.lth.se
}

\begin{document}

\maketitle

\begin{abstract}
We study the task of \emph{embodied visual active learning}, where an agent is set to explore a 3d environment with the goal to acquire visual scene understanding by actively selecting views for which to request annotation. While accurate on some benchmarks, today's deep visual recognition pipelines tend to not generalize well in certain real-world scenarios, or for unusual viewpoints. Robotic perception, in turn, requires the capability to refine the recognition capabilities for the conditions where the mobile system operates, including cluttered indoor environments or poor illumination. This motivates the proposed task, where an agent is placed in a novel environment with the objective of improving its visual recognition capability. To study embodied visual active learning, we develop a battery of agents -- both learnt and pre-specified -- and with different levels of knowledge of the environment. The agents are equipped with a semantic segmentation network and seek to acquire informative views, move and explore in order to propagate annotations in the neighbourhood of those views, then refine the underlying segmentation network by online retraining. The trainable method uses deep reinforcement learning with a reward function that balances two competing objectives: \emph{task performance, represented as visual recognition accuracy}, which requires exploring the environment, and the necessary \emph{amount of annotated data} requested during active exploration. We extensively evaluate the proposed models using the photorealistic Matterport3D simulator and show that a fully learnt method outperforms comparable pre-specified counterparts, even when requesting fewer annotations.
\end{abstract}

\section{Introduction}
Imagine a household robot in a home it has never been before and equipped with a visual sensing module to perceive its environment and localize objects. If the robot fails to recognize some objects, or to adapt to changes in the environment, over time, it may not be able to properly perform its tasks. Much of the recent success of visual perception has been achieved by deep CNNs, e.g. in image classification \cite{KrizhevskyEtAl-nips-2012, simonyan2014very, he2016deep}, semantic segmentation \cite{long2015fully, chen2017deeplab} and object detection \cite{ren2015faster, redmon2016you}. Such systems may however be challenged by unusual viewpoints or domains, as noted e.g. by \citet{ammirato2017dataset} and \citet{yang2019embodied}. Moreover, a mobile household robot should ideally operate with lightweight, re-trainable and task-specific perception models, rather than large and comprehensive ones, which could be demanding computationally and not tailored to the needs of a specific house.

\begin{figure*}[t]
     \centering
     \includegraphics[width=0.99\textwidth]{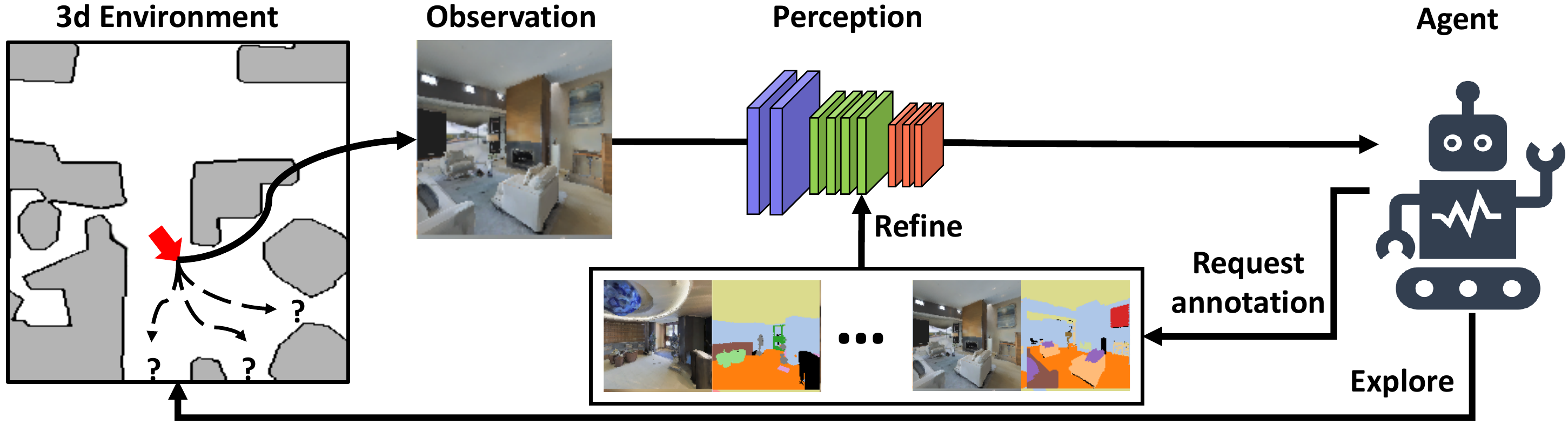}
     \caption{\textbf{Embodied visual active learning}. An agent in a 3d environment must explore and occasionally request annotation in order to efficiently refine its visual perception. The navigation component makes this task significantly more complex than traditional active learning, where the data pool over which the agent queries annotations, either in the form of image collections or pre-recorded video streams, is static and given.}
     \label{fig:first-page-fig}
\end{figure*}

In practice, even in closed but large environments, developing robust scene understanding by exhaustive approaches may be difficult, as looking everywhere requires an excessive amount of annotation labor. All views are however not equally informative, as a view containing many diverse objects is likely more useful than one covering a single semantic class, e.g. a wall. This suggests that in learning visual perception one does not have to label exhaustively. As new, potentially difficult arrangements appear in an evolving environment, it would be useful to identify those automatically, based on the task and demand, rather than programmatically, by periodically re-training a complete model. Moreover, the agent could make the most out of its embodiment by propagating a given ground truth annotation using motion -- as measured by the perceived optical flow -- in that neighborhood. The agent can then self-train, online, for increased performance. The key questions are how should one explore the environment, how to select the most informative views to annotate, and how to make the most out of them. We analyze these questions in an \emph{embodied visual active learning} framework, illustrated in \ffigure{fig:first-page-fig}. 

To ground the embodied visual active learning task, in this work we measure visual perception ability as semantic segmentation accuracy. The agent is equipped with a semantic segmentation system and must move around and request annotations in order to refine it. After exploring the scene the agent should be able to accurately segment all views in the explored area. This requires an exploration policy covering different objects from diverse viewpoints and selecting sufficiently many annotations to train the perception model. The agent can also propagate annotations to different nearby viewpoints using optical flow and then self-train. We develop a battery of methods, ranging from pre-specified ones to a fully trainable deep reinforcement learning-based agent, which we evaluate extensively in the photorealistic Matterport3D environment \cite{Matterport3D}.

\noindent In summary, our main contributions are:
\begin{itemize}
    \item We study the task of \emph{embodied visual active learning}, where an agent should explore a 3d environment to acquire visual scene understanding by actively selecting views for which to request annotation. The agent then propagates information by moving in the neighborhood of those views and self-trains;
    \item In our setup, visual learning and exploration can inform and guide one another since the recognition system is selectively and gradually refined during exploration, instead of being trained at the end of a trajectory on a full set of densely annotated views;
    \item We develop a variety of methods, both learnt and pre-specified, to tackle our task in the context of semantic segmentation;
    \item We perform extensive evaluation in a photorealistic 3d environment and show that a fully learnt method outperforms comparable pre-specified ones.
\end{itemize}

\begin{figure*}[t]
    \centering
    \includegraphics[width=0.98\textwidth]{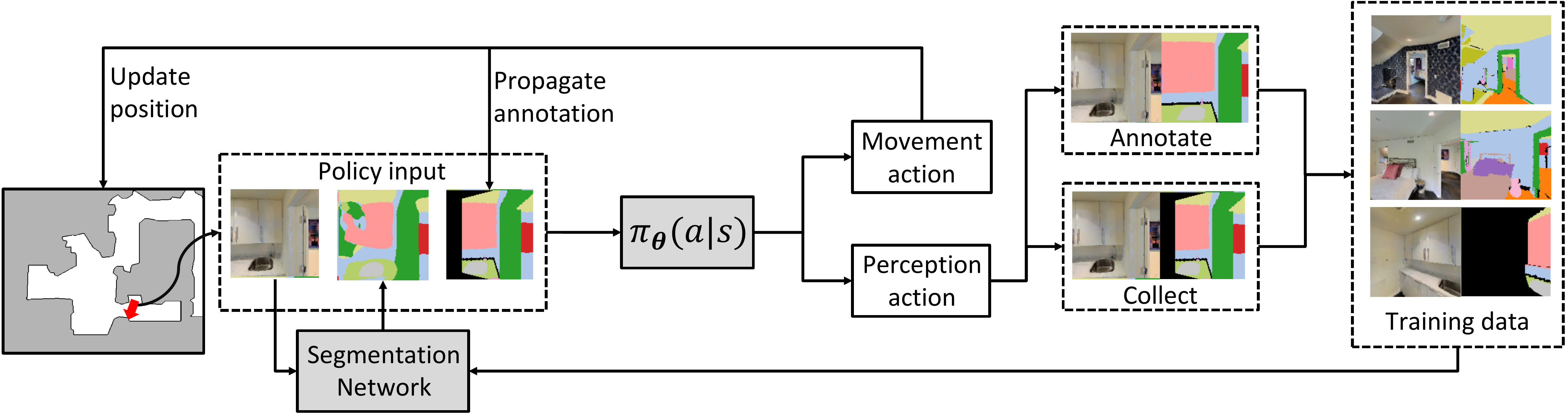}
    \caption{{\bf Embodied visual active learning for semantic segmentation}. A first-person agent is placed in a room and a deep network predicts the semantic segmentation of the agent's view. Based on the view and its segmentation, the agent can either select a \emph{movement action} to change position and viewpoint, or select a \emph{perception action} (\texttt{Annotate} or \texttt{Collect}). \texttt{Annotate} adds the current view and its ground truth segmentation to the pool of training data for the segmentation network, while \texttt{Collect} is a cheaper version (no additional supervision required) where the current view and the last annotated view -- propagated to the agent's current position using optical flow -- is added to the training set. The propagated annotation is also a policy input for the learnt agent in \S\ref{sec:real}. After a perception action, the segmentation network is refined on the current training set. The embodied visual active learning process is considered successful if, after selecting a limited number of \texttt{Annotate} actions or an exploration budget is exhausted, the segmentation network can accurately segment any other view in the environment where the agent operates. Note that the map (left) is \emph{not} provided as input to the learnt agent in \S\ref{sec:real}.}
    \label{fig:system_figure}
\end{figure*}

\section{Related Work}
The embodied visual active learning setup leverages several computer vision and machine learning concepts, such as embodied navigation, active learning and active vision. There is substantial recent literature on embodied agents navigating in real or simulated 3d environments, especially given the recent emergence of large-scale simulators \cite{savva2019habitat,ai2thor,xiazamirhe2018gibsonenv,Dosovitskiy17,savva2017minos}. 

We here briefly review variants of embodied learning. In Embodied Question Answering \cite{das2018embodied,eqa_matterport,eqa_multitarget}, an agent is given a question, e.g. "What color is the car?". The agent must typically explore the environment quite extensively in order to be able to answer. \citet{zhu2017target,mousavian2019visual} task the agent with reaching a target view using as few steps as possible. The agent receives the current view and the target as inputs in each step. In point-goal navigation \cite{mishkin2019benchmarking,midLevelReps2018,savva2019habitat,gupta2017cognitive} the agent is given coordinates of a target to reach using visual information and ego-motion. In visual exploration \cite{ramakrishnan2020exploration,Fang_2019_CVPR,chen2019learning,qi2019learning,zheng2019active, chaplot2020learning} the task is to explore an unknown environment as quickly as possible, by covering the whole scene area. In \citet{ammirato2017dataset,yang2019embodied} an agent is tasked to navigate an environment to increase the accuracy of a pre-trained recognition model, e.g. by moving around occluded objects. This is in contrast to our work where the goal is to collect views for \emph{training} a perception model. Whereas in \citet{ammirato2017dataset,yang2019embodied} the agent is spawned close to the target object, we cannot make such assumptions, as our task is not only to accurately recognize a single object or view, but to do so for \emph{all} views in the potentially large area explored by the agent.

There are relations to curiosity-driven learning \citep{pathak2017curiosity, yang2019never}, in that we also seek an agent which visits novel views (states). In \citet{pathak2017curiosity}, exploration is aided by giving rewards based on the prediction error of a self-supervised inverse-dynamics model. This is a task-independent exploration strategy useful to search 2d or 3d environments during training. In our setup, exploration is task-specific in that it is aimed specifically at refining a visual recognition system in a novel environment. Moreover, we use semi-dense rewards for both visual learning and for exploration. Hence we are not operating using sparse rewards where curiosity approaches often outperform other methods.

Our work is also related to \citet{song2015robot,pot2018self,zhong2018detect, wang2019unified}. Differently from us, \citet{song2015robot} uses hand-crafted annotation and exploration strategies, aiming to label all voxels in a 3d reconstruction by selecting a subset of frames covering all voxels. This is a form of exhaustive annotation and a visual perception system is not trained. Hence the system can only analyze objects in annotated voxels. In our setup the agent is instead tasked with both exploration and the selection of views to annotate, and we learn a perception module aiming to generalize to unseen views. In contrast to us, \citet{pot2018self, zhong2018detect, wang2019unified} do not consider an agent choosing where to move in the environment, nor which parts to label. Instead, they use all views seen when following a pre-specified path for training a visual recognition system. \citet{pot2018self} use an object detector obtained by self-supervised learning and clustering. \citet{zhong2018detect, wang2019unified} use constraints from SLAM to improve a given segmentation model. This approach could in principle complement our label propagation, and is orthogonal to our main proposals.

Next-best-view (NBV) prediction \cite{jayaraman2018learning,xiong2018snap,johns2016pairwise,jayaraman2016look,song2018im2pano3d, gartner2020deep} is superficially similar to our task. In \citet{jayaraman2018learning} an agent is trained to reveal parts of a panorama and a model is built to complete all views of the panorama. Our setup allows free movement in an environment, hence it features a navigation component which makes our task more comprehensive. While NBV typically integrates information from all predicted views, our task requires the adaptive selection of only a subset of the views encountered during the agent's navigation trajectory.

Active learning \cite{settles2009active,fang-etal-2017-learning,lughofer2012single,woodward2017active,pardo2019baod,8814236}
can be seen as the static version of our setup, as it considers approaches for learning what parts of a larger \emph{pre-existing} and \emph{static} training set should be fed into the training procedure, and in what order. We instead consider the active learning problem in an embodied setup, where an agent can move and actively select views for which to request annotation. Embodiment makes it possible to use motion to propagate annotations, hence effectively generate new ones at no additional annotation cost. In essence, our work lays groundwork towards marrying the active vision and the active learning paradigms. 

\section{Embodied Visual Active Learning}\label{sec:task-description}
Embodied visual active learning is an interplay between a first-person agent, a 3d environment and a trainable perception module. See \ffigure{fig:first-page-fig} for a high-level abstraction and \ffigure{fig:system_figure} for details of the particular task considered in this paper. The perception module processes images (views) observed by the agent in the environment. The agent can request annotations for views in order to refine the perception module. It should ideally request very few annotations as these are costly. The agent can also generate more annotations for free by neighborhood exploration using label propagation, such that when trained on that data the perception module becomes increasingly more accurate in the explored environment. To assess how successful an agent is on the task, we test how accurate the perception module is on multiple random viewpoints selected uniformly in the area explored by the agent. \\ \\
\noindent\textbf{Task overview.} The agent begins each episode randomly positioned and rotated in a 3d environment, with a randomly initialized semantic segmentation network. The ground truth segmentation mask for the first view is given for the initial training of the segmentation network. The agent can choose \emph{movement actions} (\verb|MoveForward|, \verb|MoveLeft|, \verb|MoveRight|, \verb|RotateLeft|, \verb|RotateRight| with 25 cm movements and 15 degree rotations), or \emph{perception actions}  (\verb|Annotate|, \verb|Collect|). If the agent moves or rotates, the ground truth mask is propagated using optical flow. At any time, the agent may choose to insert the propagated annotation into its training set with the \verb|Collect| action, or to ask for a new ground truth mask with the \verb|Annotate| action. After an \verb|Annotate| action the propagated annotation mask is re-initialized to the ground truth annotation. After each perception action, the segmentation network $\mathcal{S}$ is refined on the training set, which is expanded with the new data point. 

The agent's performance is evaluated at the end of the episode. The goal is to maximize the mIoU and mean accuracy of the segmentation network on the views in the area explored by the agent. Specifically, a set of \emph{reference views} are randomly sampled within a disc of radius $r$ centered at the starting location, and the segmentation network is evaluated on these. Hence to perform well the agent is required to explore its surroundings, and it should refine its perception module  in regions of high uncertainty.

\subsection{Methods for the Proposed Task}\label{sec:methods-eal}
\begin{figure}[t]
     \centering
     \includegraphics[width=0.44\textwidth]{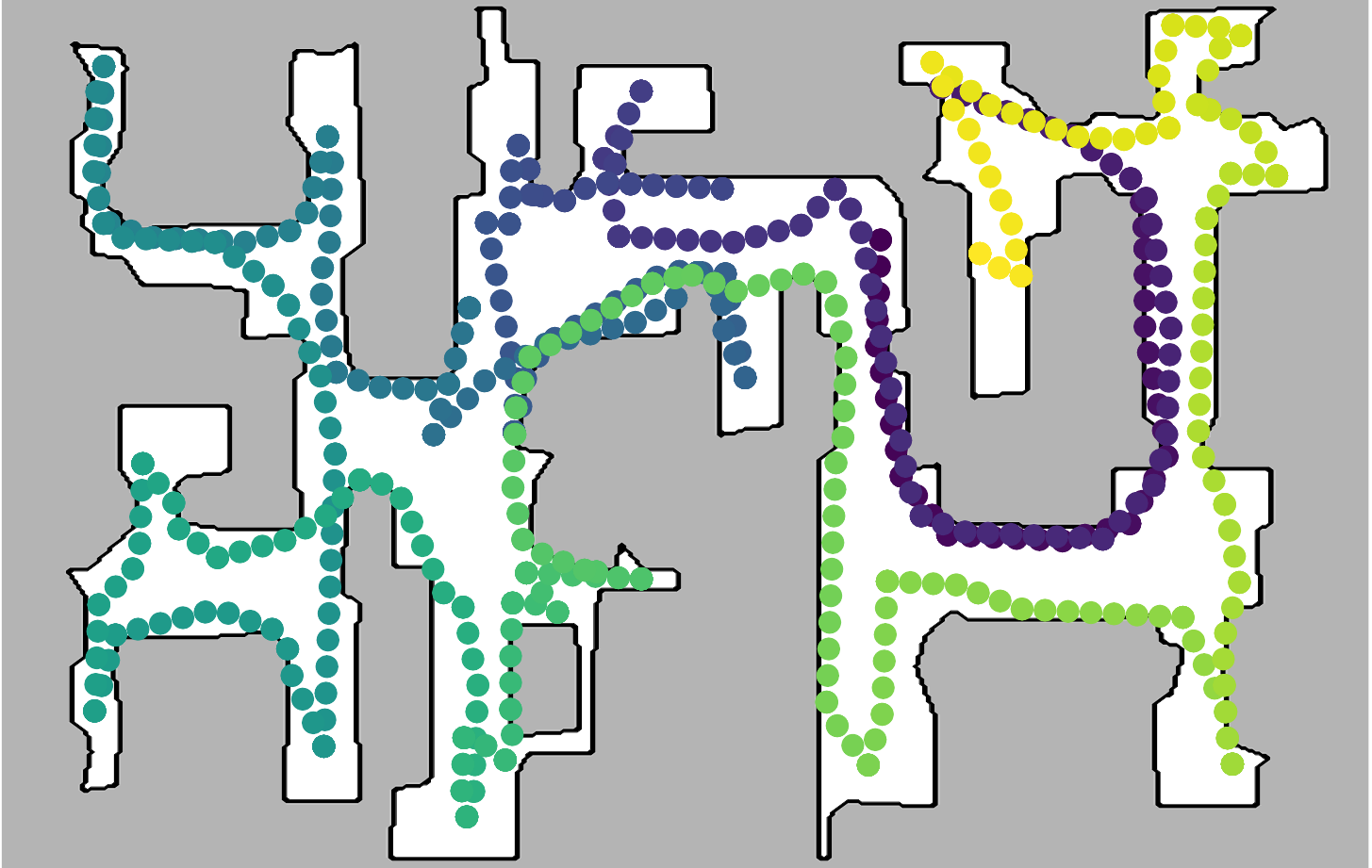}
     \caption{An example of a space filling curve in a Matterport3D floor plan. Methods based on the space filler assume complete spatial knowledge of the environment.}
     \label{fig:space-filling}
 \end{figure}

We develop several methods to evaluate and study the embodied visual active learning task. All methods except the RL-agent issue the \verb|Collect| action when 30\% of the propagated labels are unknown and \verb|Annotate| when 85\% are unknown. The intuition is that the pre-specified methods should request annotation when most pixels are unlabeled. The specific hyperparameters of all models were set based on a validation set. \\ \\
\noindent\textbf{Random.} Uniformly selects random movement actions. This baseline is thus a lower bound in terms of embodied exploration for this task. \\ \\
\noindent\textbf{Rotate.} Continually rotates left. This method is useful in comparing with trainable agents that move and explore, i.e. to monitor what improvements can be expected from embodiment. \\ \\
\noindent\textbf{Bounce.} Explores by walking straight forward until it hits a wall, then samples a new random direction and moves forward until it collides with a new wall, and so on. This agent quickly explores the environment. \\ \\
\noindent\textbf{Frontier exploration.} This method builds a map, online, by using using depth and motion from the simulator \cite{yamauchi1997frontier}. All pixels with depth within a 4m threshold are back-projected in 3d and then classified as either obstacles or navigable, based on height relative to the ground plane. This agent is confined to move within the reference view radius $r$, which is a choice to its  advantage\footnote{This ensures it is evaluated under ideal conditions in contrast to the RL-agent in \Section{sec:real}.} as annotated views will more likely be similar to reference views that reside within that same radius. \\ \\
\noindent\textbf{Space filler.} Follows a shortest space filling curve within the reference view radius $r$, and as $r$ increases the entire environment is explored. This baseline makes strong and somewhat less general (or depending on the application, altogether unrealistic) assumptions in order to create a path: knowing the floor plan in advance, as well as which locations are reachable from the start. It also only moves within the reference view radius, and knows the shortest geodesic paths to take on the curve. Hence, this method can be considered an upper bound for other methods. The space filling curve is computed by placing a grid of nodes onto the floor plan (1m resolution, using a sampling and reachability heuristic), and then finding the shortest path around it with an approximate traveling salesman solver. \Figure{fig:space-filling} shows a space filling curve in a Matterport3D floor plan. \\ \\
\noindent\textbf{RL-agent.} This fully trainable method we develop jointly learns exploration and perception actions in a reinforcement learning framework. See the full description in \S\ref{sec:real}.

\subsection{Semantic Segmentation Network}
Each method uses the same FCN-inspired deep network \cite{long2015fully} for semantic segmentation. The network consists of 3 blocks of convolutional layers, each containing 3 convolutional layers with kernels of size $3 \times 3$. The first convolutional layer in each block uses a stride of 2, which halves the resolution. For each block the number of channels doubles, using 64, 128 and 256 channels respectively. Multiple predictions are made using the final convolutional layers of each block. The multi-scale predictions are resized to the original image resolution using bilinear interpolation and are finally summed up, resulting in the final segmentation estimate. Note that we have deliberately chosen to make the network small so that it can be efficiently refined on new data.

At the beginning of each episode, the parameters are initialized randomly, and we train the network on the very first view, for which we always supply the ground truth segmentation. Each time \verb|Annotate| or \verb|Collect| is selected, we refine the network. Mini-batches of size 8, which always include the latest added labeled image, are used in training. We use random cropping and scaling for data augmentation. The network is refined either until it has trained for $1,000$ iterations or until the accuracy of a mini-batch exceeds $95\%$. We use a standard cross-entropy loss averaged over all pixels. The segmentation network is trained using stochastic gradient descent with learning rate $0.01$, weight decay $10^{-5}$ and momentum $0.9$. To propagate semantic labels, we compute optical flow between consecutive viewpoints using PWC-Net \cite{sun2018pwc}. The optical flow is computed bidirectionally and only pixels where the difference between the forward and backward displacements is less than 2 pixels are propagated \cite{sundaram2010dense}. We found that labels were reliably tracked over several frames when using 2 pixels as a threshold.

\subsection{Reinforcement Learning Agent}\label{sec:real}
To present the reinforcement-learning agent for our task, we begin with an explanation of the state-action representation and policy network, followed by the reward structure and finally policy training. 
\\ \\
\noindent\textbf{Actions, states and policy.} The agent is represented as a deep stochastic policy $\pith(a_t|s_t)$ that samples an action $a_t$ in state $s_t$ at time $t$. The actions are \verb|MoveForward|, \verb|MoveLeft|, \verb|MoveRight|, \verb|RotateLeft|, \verb|RotateRight|, \verb|Annotate| and \verb|Collect|. The full state is $s_t=\{\bs{I}_t, \bs{S}_t, \bP_t, \bs{F}_t\}$ where $\bI_t \in \mathbb{R}^{127 \times 127 \times 3}$ is the image, $\bS_t=\mathcal{S}_t(\bI_t) \in \mathbb{R}^{127 \times 127 \times 3}$ is the semantic segmentation mask predicted by the deep network $\mathcal{S}_t$ (this network is refined over an episode; $t$ indexes the network parameters at time $t$), $\bP_t \in \mathbb{R}^{127 \times 127 \times 3}$ is the propagated annotation, and $\bs{F}_t \in \mathbb{R}^{7 \times 7 \times 2048}$ is a deep representation of $\bs{I}_t$ (a ResNet-50 backbone feature map).

The policy consists of a \emph{base processor}, a \emph{recurrent module} and a \emph{policy head}. The base processor consists of two learnable components: $\phi_{img}$ and $\phi_{res}$. The 4-layer convolutional network $\phi_{img}$ takes as input the depth-wise concatenated triplet $\{\bs{I}_t, \bs{S}_t, \bs{P}_t\}$, producing $\phi_{img}(\bs{I}_t, \bs{S}_t, \bP_t) \in \mathbb{R}^{512}$. Similarly, the 2-layer convolutional network $\phi_{res}$ yields an embedding $\phi_{res}(\bs{F}_t) \in \mathbb{R}^{512}$ of the ResNet features $\bF_t$. An LSTM \cite{hochreiter1997long} with $256$ cells constitutes the recurrent module, which takes as input $\phi_{img}(\bs{I}_t, \bs{S}_t, \bP_t)$ and $\phi_{res}(\bs{F}_t)$. The input has length $1024$. The hidden LSTM state is fed to the policy head, consisting of a fully-connected layer followed by a 7-way softmax which produces action probabilities. \\ \\
\noindent\textbf{Rewards.} In training, the main reward is related to the mIoU improvement of the final segmentation network $\mathcal{S}_T$ over the initial $\mathcal{S}_0$ on a reference set $\mathcal{R}$. The set $\mathcal{R}$ is constructed at the beginning of each episode by randomly selecting views within a geodesic distance $r$ from the agent's starting location, and contains views with corresponding ground truth semantic segmentation masks. At the end of an episode of length $T$, the underlying perception module is evaluated on $\mathcal{R}$. Specifically, after an episode (with $T$ steps), the agent receives as final reward:
\begin{equation}\label{eq:reward_final}
    R_T = \miou(\mathcal{S}_T, \mathcal{R}) - \miou(\mathcal{S}_0, \mathcal{R})
\end{equation}
To obtain a denser signal, tightly coupled with the final objective, we also give a reward proportional to the improvement of $\mathcal{S}$ on the reference set $\mathcal{R}$ after each \verb|Annotate| (\emph{ann}) and \verb|Collect| (\emph{col}) action:
\begin{equation}\label{eq:reward_annot}
    R^{ann}_t = \miou(\mathcal{S}_{t}, \mathcal{R}) - \miou(\mathcal{S}_{t-1}, \mathcal{R}) - \epsilon^{ann}
\end{equation}
\begin{equation}\label{eq:reward_collect}
    R^{col}_t = \miou(\mathcal{S}_{t}, \mathcal{R}) - \miou(\mathcal{S}_{t-1}, \mathcal{R})
\end{equation}
To ensure the agent does not request costly annotations too frequently, each \verb|Annotate| action is penalized with a negative reward $-\epsilon^{ann}$ (we set $\epsilon^{ann}=0.01$), as seen in \eqref{eq:reward_annot}. Such a penalty is \emph{not} given for the free \verb|Collect| action. Moreover, the dataset we use has 40 different semantic classes, but some are very rare and apply only to small objects, and some might not even be present in certain houses. We address this imbalance by computing the mIoU using only the 10 largest classes, ranked by the number of pixels in the set of reference views for the current episode.

While the rewards \eqref{eq:reward_final} - \eqref{eq:reward_collect} should implicitly encourage the agent to explore the environment in order to request annotations for distinct, informative views, we empirically found useful to include an additional explicit exploration reward. Denote by $\{\bx_i\}_{i=1}^{t-1} = \{(x_i,y_i)\}_{i=1}^{t-1}$ the positions the agent has visited up to time $t-1$ in its current episode, and let $\bx_t=(x_t,y_t)$ denote its current position. We define the exploration (\emph{exp}) reward based on a kernel density estimate of the agent's visited locations:
\begin{equation}\label{eq:reward_exploration}
R_t^{exp} = a-bp_t(\bx_t) \coloneqq a - \frac{b}{t-1} \sum_{i=1}^{t-1} k(\bx, \bx_i) 
\end{equation}
where $a$ and $b$ are hyperparameters (both set to 0.003). Here $p_t(\bx_t)$ is a Gaussian kernel estimate of the density with bandwidth 0.3m. It is large for previously visited positions and small for unvisited positions, thereby encouraging the agent's expansion towards new places in the environment. The exploration reward is only given for movement actions. Note that the pose $\bx_i$ is only used to compute the reward $R_t^{exp}$ and is not available to the policy via the state space.  \\ \\
\textbf{Policy training.} The policy network is trained using PPO \cite{schulman2017proximal} based on the RLlib reinforcement learning package \cite{liang2018rllib}, as well as OpenAI Gym \cite{brockman2016openai}. For optimization we use Adam \cite{kingma2014adam} with batch size 512, learning rate $10^{-4}$ and discount rate $0.99$. During training, each episode consists of $256$ actions. The agent is trained for 4k episodes, which totals 1024k steps.

Our system is implemented in TensorFlow \cite{abadi2016tensorflow}, and it takes about 3 days to train an agent using 4 Nvidia Titan X GPUs. An episode of length 256 took on average about 3 minutes using a single GPU, and during training we used 4 workers with one GPU each, collecting rollouts independently. The runtime per episode varies depending on how frequently the agent decides to annotate, as training the segmentation network is the bottleneck and accounts for approximately 90\% of the run-time. We used optical flow from the simulator to speed up policy training. For evaluation, the RL-agent and all other methods use PWC-Net to compute optical flow. The ResNet-50 feature extractor is pre-trained on ImageNet \cite{JiaDeng2009imagenet} with weights frozen during policy training. 

\begin{figure*}[!htbp]
    \centering
    \includegraphics[width=0.497\textwidth]{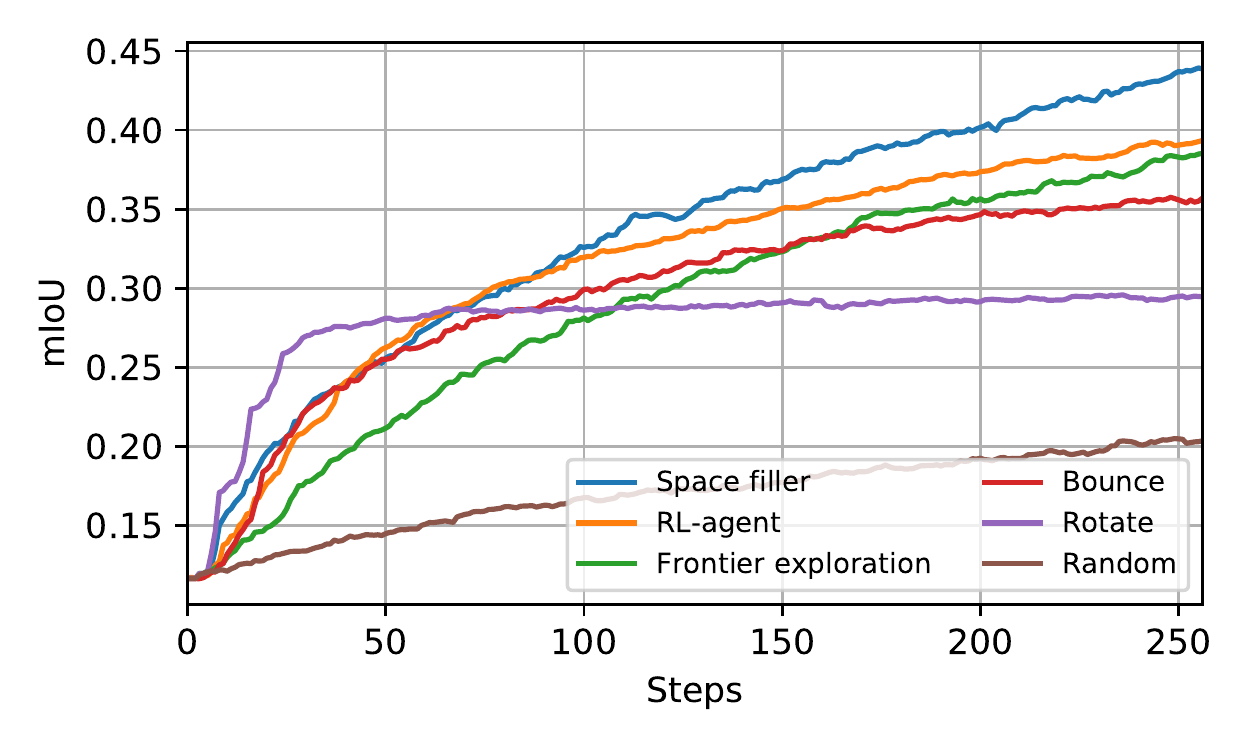}
    \includegraphics[width=0.497\textwidth]{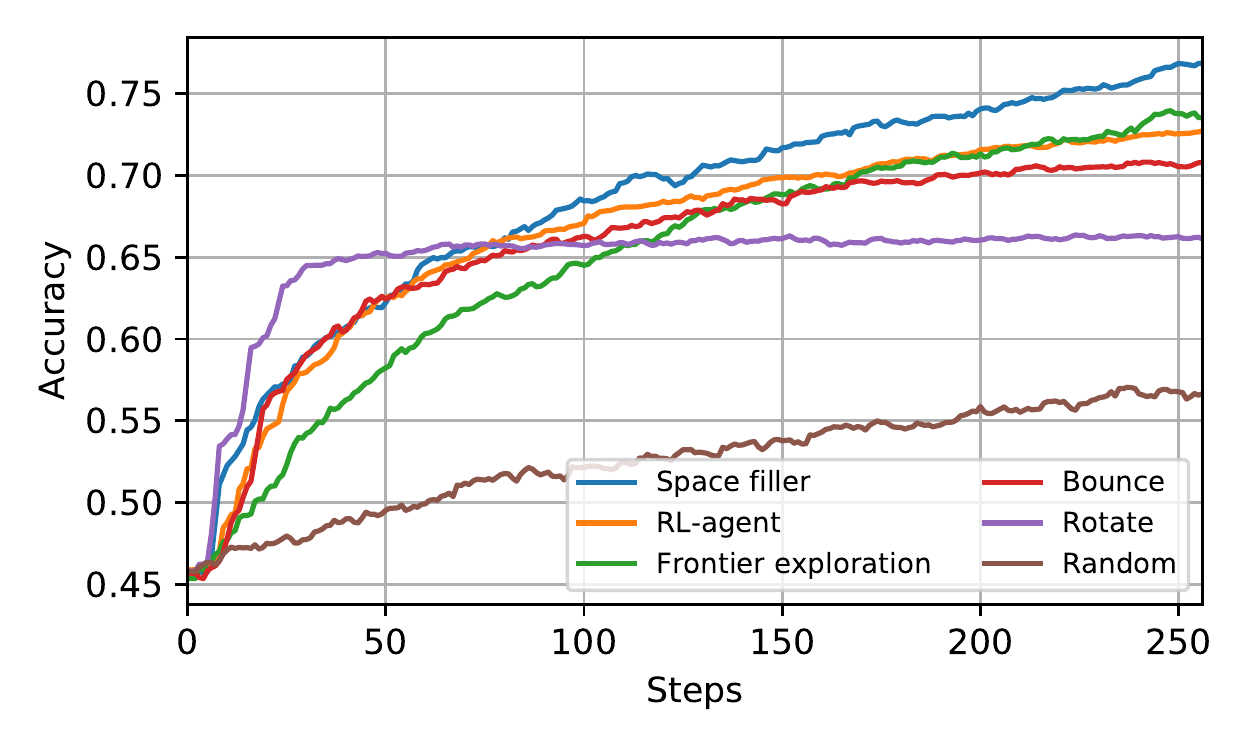}
    \caption{Mean segmentation accuracy and mIoU versus number of actions (steps), evaluated on the Matterport3D test scenes. The RL-agent was trained on 256-step episodes. This agent fairly quickly outperforms all other comparable pre-specified agents. \emph{Rotate} is strong initially since it quickly gathers many annotations in a 360 degree arc, but is eventually outperformed by most other methods that move around in the houses. Frontier exploration yields similar accuracy as the RL-agent after about 170 steps, but uses significantly more annotations (cf. \tableref{table:table1}) and assumes perfect pose and depth information. The space filler, which assumes full knowledge of the environment, yields the best results after about 100 steps.}
    \label{fig:acc_vs_steps_testing}
\end{figure*}

\section{Experiments}\label{sec:experiments}
In this section we provide empirical evaluations of various methods. The primary metrics are mIoU and segmentation accuracy but we emphasize that we test the exploration and annotation selection capability of \emph{policies} -- the mIoU and accuracy measure how well agents explore in order to refine their perception. Differently from accuracy, the mIoU does not become overly high by simply segmenting large background regions (such as walls), hence it is more representative of overall semantic segmentation quality. \\ \\
\noindent\textbf{Experimental setup.} We evaluate the methods on the Matterport3D dataset \cite{Matterport3D} using the embodied agent framework Habitat \cite{savva2019habitat}. This setup allows the agent to freely explore photorealistic 3d models of large houses, that have ground truth annotations for 40 diverse semantic classes. Hence it is a suitable environment for evaluation. To assess the generalization capability of the RL-agent we train and test it in \emph{different} houses. We use the same 61, 11 and 18 houses for training, validation and testing as \citet{Matterport3D}. The RL-agent and all pre-specified methods except the space filler are \emph{comparable} in terms of assumptions, cf. \Section{sec:methods-eal}. The space filler assumes full spatial knowledge of the environment (ground truth map) and hence has inherent advantages over the other methods.

During RL-agent training we randomly sample starting positions and rotations from the training houses at the start of each episode. An episode ends after $256$ actions. Hyperparameters of the learnt and pre-specified agents are tuned on the validation set. For validation and testing we use 3 and 4 starting positions per scene, respectively, so each agent is tested for a total of 33 episodes in validation and 72 episodes in testing. The reported metrics are the mean over all these runs. All methods are evaluated on the same starting positions in the same houses. The reference views used to evaluate the semantic segmentation performance are obtained by sampling $32$ random views within a $5$ m geodesic distance of the agent's starting position at the beginning of each episode. In training the reference views are sampled randomly. During validation and testing, for fairness, we sample the same views for a given starting position when we test different agents. Note that there is no overlap between reference views during policy training and testing, since training, validation and testing houses are non-overlapping.

\begin{figure*}[!htbp]
    \centering
    \includegraphics[width=0.497\textwidth]{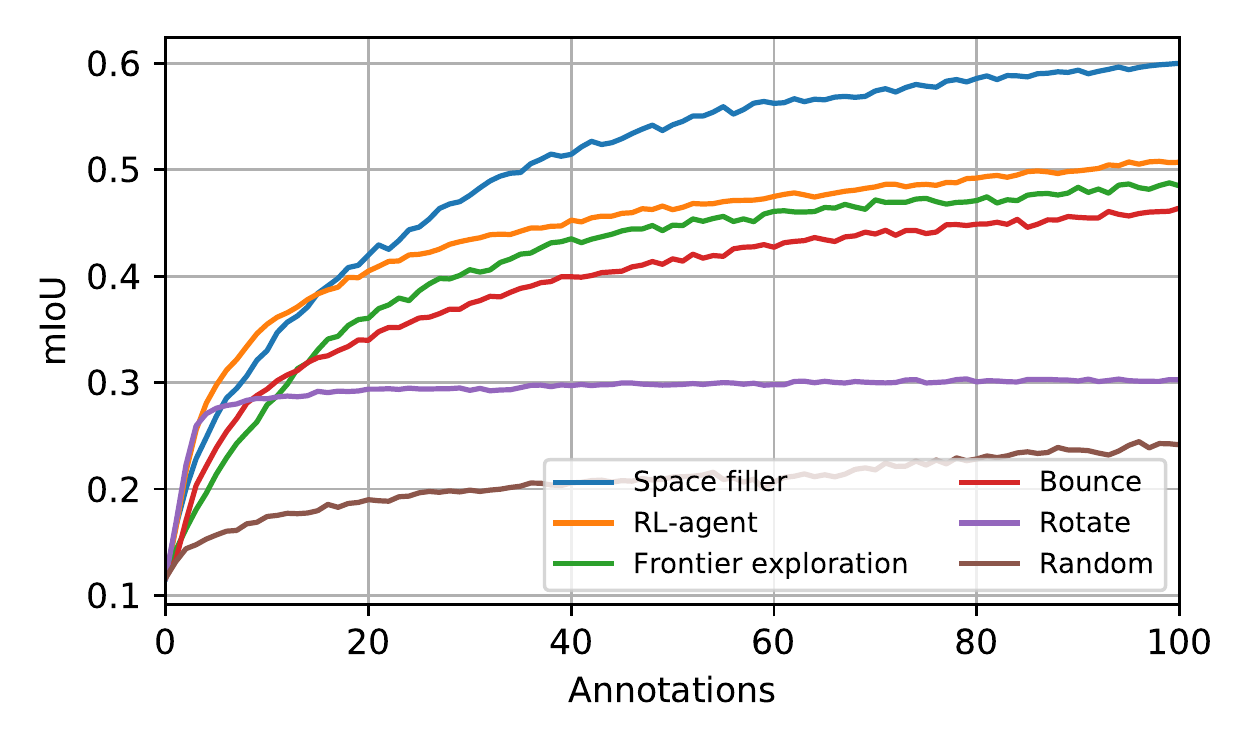}
    \includegraphics[width=0.497\textwidth]{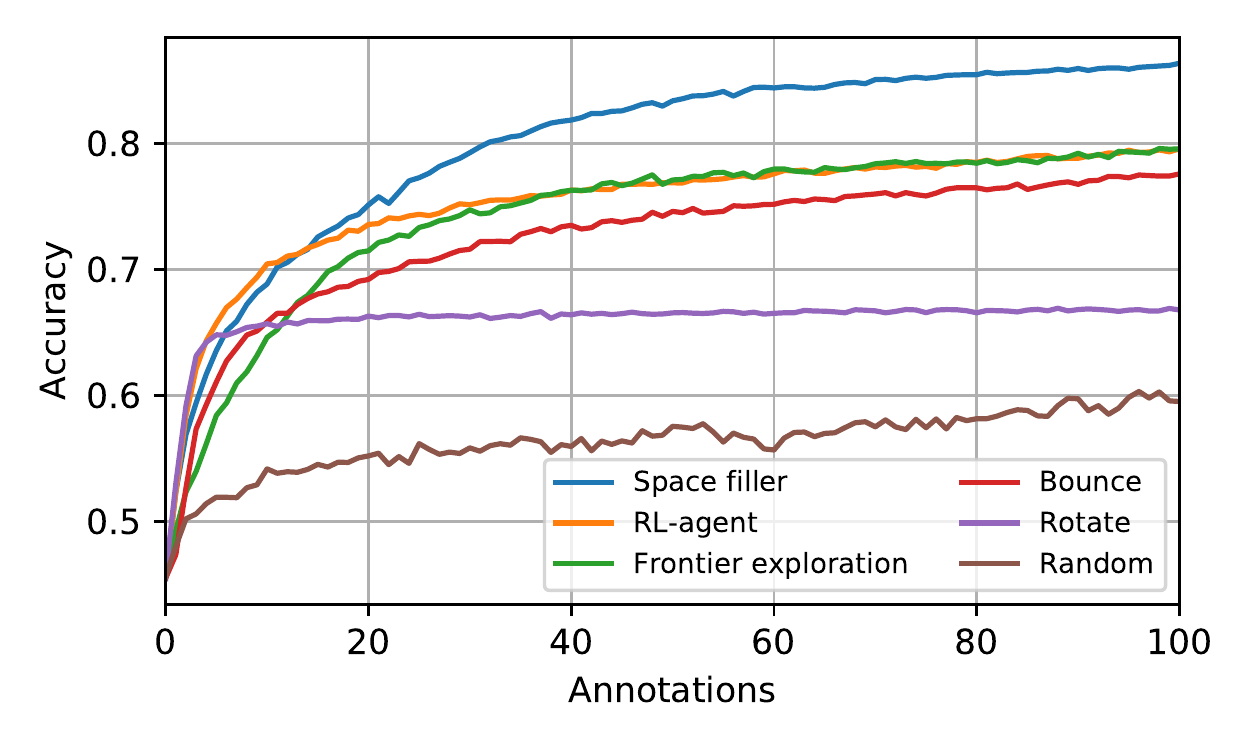}
    \caption{Mean segmentation accuracy and mIoU for a varying number of requested annotations evaluated on the Matterport3D test scenes. The RL-agent outperforms all comparable pre-specified methods (although frontier exploration matches it in accuracy after about 40 annotations), indicating that it has learnt an exploration policy which generalizes to novel scenes. The space filler, as expected, outperforms the RL-agent, except for less than 15 annotations. Thus the RL-agent is best before and around its training regime, where on average annotates 16.7 times per episode, cf. \tableref{tab:fixed_ep_length}.}
    \label{fig:acc_vs_annots_testing}
\end{figure*}

Recall that the RL-agent's policy parameters are denoted by $\bth$. Let $\bth_{seg}$ denote the parameters of the underlying semantic segmentation network, in order to clarify when we reset, freeze and refine $\bth$ and $\bth_{seg}$, respectively. For RL-training, we refine $\bth$ during policy estimation in the training houses. When we evaluate the policy on the validation or test houses we freeze $\bth$ and only use the policy for inference. The parameters of the segmentation network $\bth_{seg}$ are always reset at the beginning of an episode, regardless of which house we deploy the agent in, and regardless of whether the policy network is training or not. During an episode, we refine $\bth_{seg}$ exactly when the agent selects the \verb|Annotate| or \verb|Collect| actions (this applies also to all the other methods described in \Section{sec:methods-eal}). Thus annotated views in an episode are used to refine $\bth_{seg}$ in that episode only, and are not used in any other episodes.

\begin{table}[t]
\caption{Comparison of different agents for a fixed episode length of 256 actions on the Matterport3D test scenes. The RL-agent gets higher mIoU using far fewer annotations than comparable pre-specified methods, implying that the RL-agent's policy selects more informative views to annotate.}\label{table:table1}
    \centering\begin{tabular}{c@{\hskip 4mm} c@{\hskip 4mm} c@{\hskip 4mm} c@{\hskip 4mm} c}\hline
        Method & mIoU & Acc & \# Ann & \# Coll \\ \hline
        Space filler & $0.439$ & $0.769$ & $24.7$ & $23.9$ \\
        RL-agent & $0.394$ & $0.727$ & $16.7$ & $ 15.2 $ \\
        Frontier exploration & $ 0.385 $ & $ 0.735 $ & $ 24.2 $ & $ 21.6 $ \\
        Bounce & $ 0.357 $ & $ 0.708$ & $ 29.6$ & $ 26.0 $ \\
        Rotate & $  0.295 $ &  $ 0.661 $ & $ 34.3 $ & $ 32.7 $ \\
        Random & $ 0.204 $ & $ 0.566$  & $ 29.1 $ & $ 19.5 $ \\ \hline
    \end{tabular}
    \label{tab:fixed_ep_length}
\end{table}
\begin{table}[t]
\caption{Comparison of different agents for a fixed budget of 100 annotations on Matterport3D test scenes. The RL-agent gets a higher mIoU than comparable pre-specified agents, despite not being trained in this setting.}
\centering\begin{tabular}{c@{\hskip 4mm} c@{\hskip 4mm} c@{\hskip 4mm} c@{\hskip 4mm} c}\hline
    Method & mIoU & Acc & \# Steps & \# Coll \\ \hline
    Space filler & $ 0.600 $ & $ 0.863 $ & $ 1048 $ & $ 91 $ \\
    RL-agent & $ 0.507 $ & $ 0.796 $ & $ 1541 $ & $ 94 $ \\ 
    Frontier exploration & $ 0.485 $ & $ 0.796 $ & $ 998 $ & $ 84 $ \\
    Bounce & $ 0.464 $ & $ 0.776 $ & $861$ & $87$ \\
    Rotate & $ 0.303 $ & $ 0.668 $ & $ 752 $ & $ 96 $ \\
    Random & $ 0.242 $ & $ 0.595 $ & $ 910 $ & $ 64 $ \\ \hline
\end{tabular}
\label{tab:fixed_budget}
\end{table}

\subsection{Main Results}\label{sec:main-results}
\begin{figure*}[!htbp]
    \centering
    \includegraphics[width=0.97\textwidth]{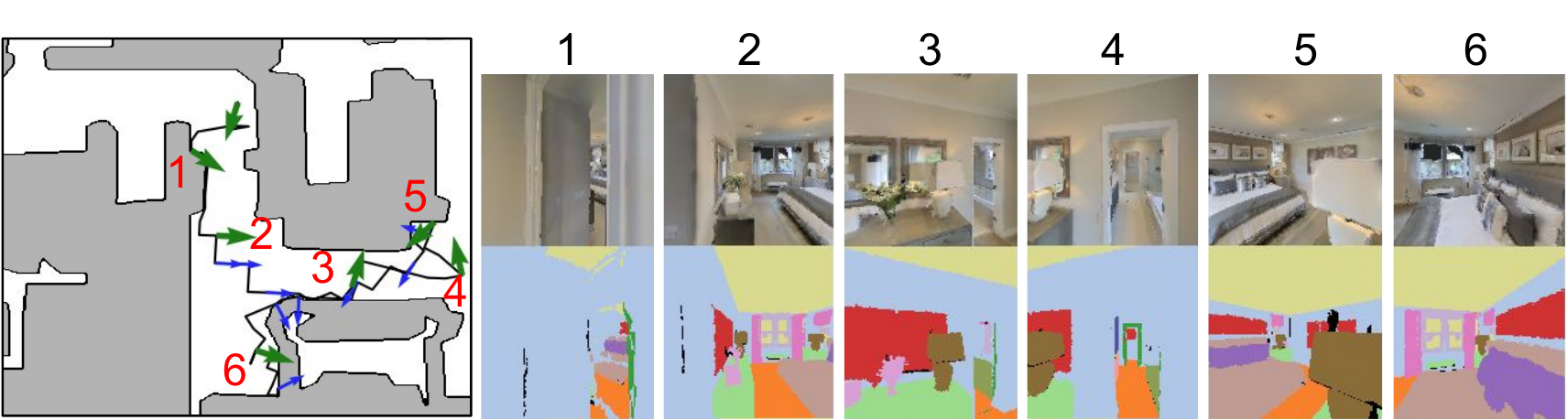}
    \caption{The first six requested annotations by the RL-agent in a room from the test set. Left: Map showing the agent's trajectory and the six first requested annotations (green arrows). The initially given annotation is not indicated with a number. Blue arrows indicate \texttt{Collect} actions. Right: For each annotation (numbered 1 - 6) the figures show the image seen by the agent and the ground truth received when the agent requested annotations. As can be seen, the agent quickly explores the room and requests annotations containing diverse semantic classes.}
    \label{fig:qualitative_examples}
\end{figure*}

\begin{figure*}[!htbp]
    \centering
    \includegraphics[width=0.99\textwidth]{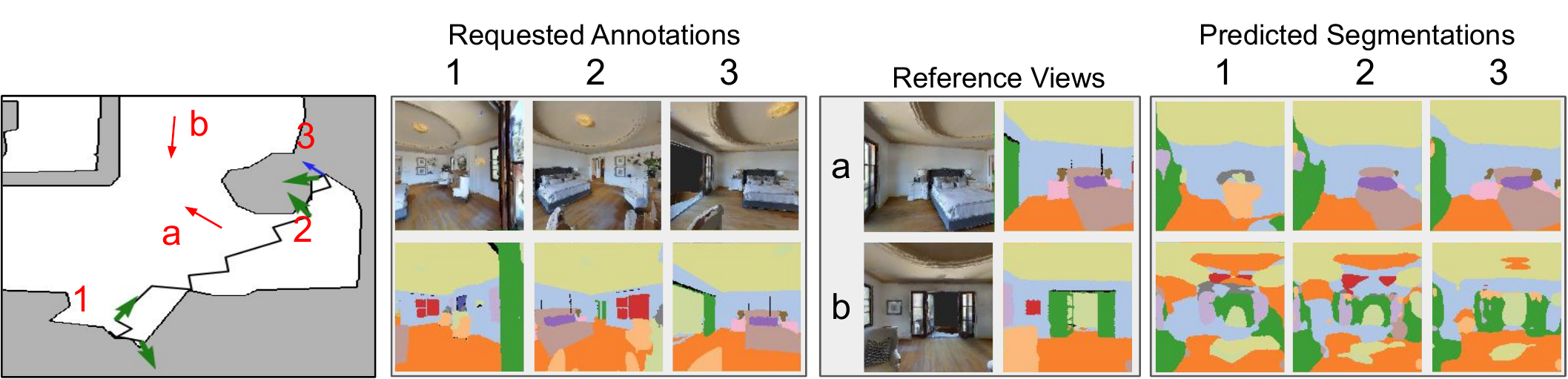}
    \caption{Example of the RL-agent's viewpoint selection and how its perception improves over time. We show results of two reference views after the first three annotations of the RL-agent. Left: Agent's movement path is drawn in black on the map. The annotations (green arrows) are numbered 1 - 3, and the associated views are shown immediately right of the map (the initially given annotation is not shown). Red arrows labeled \textit{a} - \textit{b} indicate the reference views. Right: Reference views and ground truth masks, followed by predicted segmentation after one, two and three annotations. Notice clear segmentation improvements as the agent requests more annotations. Specifically, note how reference view \textit{a} improves drastically with annotation \textit{2} as the bed is visible in that view, and with annotation \textit{3} where the drawer is seen. Also note how segmentation improves for reference view \textit{b} after the door is seen in annotation \textit{3}.}
    \label{fig:qualitative_examples2}
\end{figure*}

We measure the performances of the agents in two settings: (a) with unlimited annotations but limited total actions (max 256, as during RL-training), or (b) for a limited annotation budget (max 100) but unlimited total actions. All methods were tuned on the validation set in a setup similar to (a) with 256 step episodes. Note however that the number of annotations can differ for different methods in a 256 step episode. The setup (b) is used to assess how the different methods compare for a fix number of annotations. \\
\noindent\textbf{Fixed episode length.} \Table{tab:fixed_ep_length} and \ffigure{fig:acc_vs_steps_testing} show results on the test scenes for episodes of length 256. The RL-agent outperforms the comparable pre-specified methods in mIoU and accuracy, although frontier exploration -- which uses perfect pose and depth information, and is idealized to always move within the reference view radius -- yields similar accuracy after about 170 steps. The RL-agent uses much fewer annotations than other methods, hence those annotated views are more informative. The space filler, which assumes perfect knowledge of the map, outperforms the RL-agent but uses significantly more annotations. Note that the \emph{Rotate} baseline saturates, supporting the intuition that an agent has to move around in order to increase performance in complex environments.
\\ \\
\noindent\textbf{Fixed annotation budget.} In \tableref{tab:fixed_budget} and \ffigure{fig:acc_vs_annots_testing} we show test results when the annotation budget is limited to 100 images per episode. As expected, the space filler yields the best results, although the RL-agent gets comparable performance when using up to 15 annotations. The RL-agent outperforms comparable pre-specified methods in mIoU and accuracy. Frontier exploration obtains similar accuracy. We also see that the episodes of the RL-agent are longer.
\\ \\
\noindent\textbf{Qualitative examples.}
\Figure{fig:qualitative_examples} shows examples of views that the RL-agent choose to annotate. The agent explores large parts of the space and the annotated views are diverse, both in their spatial locations and in the types of semantic classes they contain. \Figure{fig:qualitative_examples2} shows how the segmentation network's performance on two reference views improves during an episode. The two views are initially poorly segmented, but as the agent explores and acquires annotations for novel views, the accuracy on the reference views increases.

\subsection{Ablation Studies of the RL-agent}\label{sec:ablations}
Ablation results of the RL-agent on the validation set are in \tableref{tab:ablation}. We compare to the following versions: i) Policy without visual features $\phi_{img}$; ii) Policy without ResNet features $\phi_{res}$; iii) No additional exploration reward \eqref{eq:reward_exploration}, i.e. $R_t^{exp}=0$; iv) No \verb|Collect| action and $\bs{P}_t$ is not an input to $\phi_{img}$; and v) Only exploration trained, using the heuristic strategy for annotations. We trained the ablated models for 4,000 episodes as for the full model.

Both the validation accuracy and mIoU are higher for the full RL-model compared to all ablated variants, justifying design choices. The model not relying on propagating annotations and using the \verb|Collect| action performs somewhat worse than the full model despite a comparable amount of annotations. The learnt annotation strategy yields higher mIoU and accuracy compared to the heuristic one, at comparable number of annotations. The exploration reward is important in encouraging the agent to navigate to unvisited positions -- without it performance is worse, despite a comparable number of annotations. The agent trained without the exploration reward uses an excessive number of \verb|Collect| actions, so this agent often stands still instead of moving. Finally, omitting either visual or ResNet features from the policy significantly harms accuracy for the resulting recognition system. 

\begin{table}[t]
\centering
    \caption{Ablation study of different RL-based model variants for 256-step episodes on the validation set. The full RL-agent outperforms all ablated models at a comparable or lower number of requested annotations.}
    \begin{tabular}{c@{\hskip 4mm} c@{\hskip 4mm} c@{\hskip 4mm} c@{\hskip 4mm} c}\hline
        Variant & mIoU & Acc & \# Ann & \# Coll  \\ \hline
        Full model & $ 0.427 $ & $ 0.732 $ & $ 16.4 $ & $ 16.4 $ \\
        No collect nor $\bs{P}_t$ & $ 0.415 $  & $ 0.727 $  & $ 17.9 $ & $ 0.0 $ \\
        Only exploration & $ 0.411 $ & $ 0.727 $ & $ 16.1 $ & $ 14.4 $ \\
        $R_t^{exp}=0$ & $ 0.401 $ & $ 0.719 $ & $ 17.7 $ & $ 47.4 $ \\ 
        No $\phi_{img}$ & $ 0.378 $  & $ 0.696 $ & $ 14.3 $ & $ 3.8 $ \\
        No ResNet & $ 0.375 $ & $ 0.705 $ & $ 23.3 $ & $ 0.3 $ \\
         \hline
    \end{tabular}
    \label{tab:ablation}
\end{table}

\subsection{Analysis of Annotation Strategies}\label{sec:results-extra}
In this section we examine how different annotation strategies affect the task performance on the validation set for the space filler and bounce methods. 
Specifically, the annotation strategies are:
\begin{itemize}
    \item \textbf{Threshold perception.} This is the variant evaluated in \S\ref{sec:main-results}, i.e. it issues the \verb|Collect| action when 30\% of the propagated labels are unknown and \verb|Annotate| when 85\% are unknown.
    \item \textbf{Learnt perception.} We train a simplified RL-agent where the movement actions are restricted to follow the exploration trajectory of the baseline method (space filler and bounce, respectively). This model has 3 actions: move along the baseline exploration path, \verb|Annotate| and \verb|Collect|. All other training settings are identical to the full RL-agent.
    \item \textbf{Random perception.} In each step, this variant follows the baseline exploration trajectory with 80\% probability, while annotating views and collecting propagated labels with 10\% probability each.
\end{itemize}
\begin{table}[t]
\centering
\caption{Results for different model variants of the space filler method. We report the mean on the validation scenes. The threshold perception strategy -- which is the one used in the main evaluations in \Section{sec:main-results} -- yields the best results.}
    \begin{tabular}{c@{\hskip 4mm} c@{\hskip 4mm} c@{\hskip 4mm} c@{\hskip 4mm} c}\hline
        Variant & mIoU  & Acc & \# Ann & \# Coll \\ 
        \hline
        Threshold perception & $ 0.472 $ & $ 0.770 $ & $ 20.8 $ & $ 19.9 $ \\
        Learnt perception & $ 0.454 $ & $ 0.755 $ & $ 22.8 $ & $ 37.4 $ \\
        Random perception & $ 0.446 $ & $ 0.747 $ & $ 24.2 $ & $ 24.4 $ \\
         \hline
    \end{tabular}
    \label{tab:space_filler_ablation}
\end{table}
As can be seen in \tableref{tab:space_filler_ablation}, the best results for the space filler are obtained by using the threshold strategy, which also annotates slightly less frequently than other variants. Using learnt perception actions yields better results compared to random perception actions, and takes slightly fewer annotations per episode. Similar results carry over to the bounce method in \tableref{tab:bounce_ablation}, i.e. the best results are again obtained by the threshold variant. The model with a learnt annotation strategy fails to converge to anything better than heuristic perception strategies. In fact, it converges to selecting \verb|Collect| almost 40\% of the time, which indicates a lack of movement for this variant. 

In \tableref{tab:ablation} we saw that a learnt exploration method with a heuristic annotation strategy yields worse results than a fully learnt model. Conversely, the results from \tableref{tab:space_filler_ablation} and \tableref{tab:bounce_ablation} show that a heuristic exploration method using a learnt annotation strategy yields worse results than an entirely heuristic model. Together these results indicate that it is necessary to learn how to annotate and explore jointly to provide the best results, given comparable environment knowledge. 

\begin{table}[t]
\centering
\caption{Results for different model variants of the bounce method. We report the mean on the validation scenes. The threshold perception strategy -- which is the one used in the main evaluations in \Section{sec:main-results} -- yields the best results, but also uses the largest amount of annotations on average.}
    \begin{tabular}{c@{\hskip 4mm} c@{\hskip 4mm} c@{\hskip 4mm} c@{\hskip 4mm} c}\hline
        Variant & mIoU  & Acc & \# Ann & \# Coll  \\ \hline 
        Threshold perception & $ 0.388 $ & $ 0.706 $ & $ 27.4 $ & $ 24.5 $ \\ Learnt perception & $ 0.375 $ & $ 0.699 $ & $ 14.6 $ & $ 98.8 $ \\
        Random perception & $ 0.379 $ & $ 0.698 $ & $ 25.9 $ & $ 24.6 $ \\
         \hline
    \end{tabular}
    \label{tab:bounce_ablation}
\end{table}
\subsection{Pre-training the Segmentation Network}
Recall that our semantic segmentation network is randomly initialized at the beginning of each episode. In this section we evaluate the effect of instead pre-training the segmentation network\footnote{In this pre-training experiment, we use the same architecture and hyperparameters for the segmentation network as when it is trained and deployed in the embodied visual active learning task.} on the 61 training houses using about 20,000 random views. In \tableref{tab:segnet_pretraining} we compare using this pre-trained segmentation network as initialization for the RL-agent with the case of random initialization. We also show results when not further fine-tuning the pre-trained segmentation network, i.e. when not performing any embodied visual active learning.

The weak result obtained when not fine-tuning (first row) indicates significant appearance differences between the houses. This is further suggested by the fact that the RL-agent gets a surprisingly modest boost from pre-training the segmentation network (third row vs second row). Note the different number of annotated views used here --  the agent without pre-training uses only 16.4 views on average, while the other uses about $20,000 + 14.4$ annotated views, if we count all the images used for pre-training. Due to relatively marginal gains for a large number of annotated images, we decided to evaluate all agents without pre-training the segmentation network.

\begin{table}[t]
\centering
\caption{Results for different training regimes for the semantic segmentation network. A pre-trained segmentation network generalizes poorly to unseen environments (first row), and there is relatively little gain for the RL-agent by having a pre-trained segmentation network (third row). Note that pre-training uses over 1000x more annotations compared to performing embodied active visual learning from scratch.}
\begin{tabular}{c@{\hskip 4mm} c@{\hskip 4mm} c@{\hskip 4mm} c@{\hskip 4mm} c}\hline
        Variant & mIoU  & Acc & \# Ann & \# Coll  \\ \hline
        Pre-train, no RL & $ 0.208 $ & $ 0.549 $ & $ \mathrm{20k} $ & $ 0.0 $ \\
        No pre-train, RL & $ 0.427 $ & $ 0.732 $ & $ 16.4 $ & $ 16.4 $ \\
        Pre-train, RL & $ 0.461 $ & $ 0.780 $ & $ \mathrm{20k}+14.4 $ & $ 13.3 $ \\
         \hline
    \end{tabular}
    \label{tab:segnet_pretraining}
\end{table}

\section{Conclusions}
In this paper we have explored the \emph{embodied visual active learning} task for semantic segmentation and developed a diverse set of methods, both pre-designed and learning-based, in order to address it. The agents can explore a 3d environment and improve the accuracy of their semantic segmentation networks by requesting annotations for informative viewpoints, propagating annotations via optical flow at no additional cost by moving in the neighborhood of those views, and self-training. We have introduced multiple baselines as well as a more sophisticated fully learnt model, each exposing different assumptions and knowledge of the environment. Through extensive experiments in the photorealistic Matterport3D environment we have thoroughly investigated the various methods and shown that the fully learning-based method outperforms comparable non-learnt approaches, both in terms of accuracy and mIoU, while relying on fewer annotations. 

{\small \noindent{\bf Acknowledgments:} This work was supported in part by the European Research Council Consolidator grant
SEED, CNCS-UEFISCDI PN-III-P4-ID-PCE-2016-0535 and PCCF-2016-0180, the
EU Horizon 2020 Grant DE-ENIGMA, Swedish Foundation for Strategic Research (SSF) Smart Systems Program, as well as the Wallenberg AI, Autonomous Systems and Software Program (WASP) funded by the Knut and Alice Wallenberg Foundation.}

\bibliography{aaai2021_arxiv}

\end{document}